\documentclass[sigconf]{acmart}

\setcopyright{none}
\AtBeginDocument{%
  }

\acmConference[ADScAI Conference]
{Second Annual Conference of Department of Computer Science & Engineering}
{2026}
{University of Moratuwa, Sri Lanka}

\acmYear{2026}
\acmDOI{}
\acmISBN{}

\begin{document}

\title{Adversarial Stress Testing of Role-Playing Language Agents using Multi-Agent Evaluation}

\author{
\parbox{\textwidth}{
\centering
Saqib Shouqi \quad Abdullah Nazly \quad Januki Wanniarachchi \quad Ravisha De Alwis\\[4pt]
\small
School of Computing, Informatics Institute of Technology, Colombo, Sri Lanka\\[2pt]
\footnotesize
\texttt{\{saqib.20231934, abdullah.20230976, januki.20232291, ravisha.20231198\}@iit.ac.lk}
}
}

\renewcommand{\shortauthors}{Saqib Shouqi, Abdullah Nazly, Januki Wanniarachchi, Ravisha De Alwis}

\begin{abstract}
Role-Playing Language Agents (RPLAs) are increasingly deployed in high-stakes 
applications such as healthcare assistance, customer support, and education, 
where maintaining consistent personas, ethical constraints, and behavioral 
coherence under adversarial pressure is critical. Existing evaluation approaches 
rely on static benchmarks or isolated single-turn prompts that fail to capture 
cumulative behavioral failures emerging over extended interactions.

We present a modular multi-agent platform for adversarially 
stress-testing RPLAs through structured, multi-turn dialogue. The system 
coordinates three agents: a strategy-driven Interrogator Agent that applies 
six progressive adversarial strategies, a Target Agent representing the RPLA 
under evaluation, and an automated Judging Agent that scores behavior across 
role fidelity, drift, ethical deviation, and consistency dimensions.

Through experiments across three personas and three LLM families, we 
demonstrate that multi-strategy adversarial evaluation reveals failure modes 
invisible to single-strategy testing, reducing overall robustness scores by 
0.17--0.20 points on average. Cross-model validation confirms consistent 
degradation patterns across Llama-3.3-70B, GPT-4o-mini, and Claude-3.5-Haiku, 
with Authority Challenge and Emotional Manipulation emerging as the most 
effective attack strategies. Automated judging achieves strong human alignment 
($r = 0.82$, Fleiss' $\kappa = 0.71$). This work is released as an 
open-source platform to support AI safety and reproducible RPLA benchmarking. 
While the framework enables systematic discovery of failure modes, we acknowledge 
potential ethical risks associated with adversarial testing methodologies and 
emphasize responsible usage for improving AI safety.
\end{abstract}


\maketitle

\section{Introduction}

Large Language Models (LLMs) are increasingly deployed as interactive agents that simulate defined roles, such as customer support representatives, tutors, therapists, and non-player characters. In these contexts, success is not determined solely by linguistic fluency, but by an agent's ability to maintain role consistency, respect ethical constraints, and behave coherently over long, multi-turn interactions.

Despite these requirements, most current evaluation methodologies remain limited. Benchmark datasets and single-turn prompts fail to model adversarial or malicious user behavior, while human evaluation is expensive, subjective, and difficult to scale. As a result, many deployed agents exhibit failures such as role abandonment, ethical violations, or contradictory behavior when exposed to adversarial inputs.

To address this gap, we propose a modular, multi-agent framework designed to adversarially stress-test role-playing language agents in realistic interaction settings. Unlike static evaluations, the proposed framework introduces an active adversarial agent that systematically probes weaknesses, while an automated judging agent provides structured, quantitative assessments of agent behavior.

The contributions of this work are:
\begin{itemize}
  \item A production-grade, extensible platform for adversarial evaluation of RPLAs
  \item A strategy-driven agent for systematic failure discovery
  \item A multi-dimensional automated evaluation framework
  \item A reproducible experimental pipeline with persistent storage and visualization
\end{itemize}

\section{Related Work}

Prior research on LLM safety and alignment has explored issues such as harmful content generation, bias, hallucinations, and prompt injection attacks. Studies on jailbreaks and prompt injection demonstrate that LLMs can be manipulated to violate system-level constraints through carefully crafted inputs. However, most work focuses on single-turn attacks rather than sustained adversarial dialogues.

Table~\ref{tab:related_work} summarizes key differences between the proposed framework and related approaches, highlighting its focus on multi-turn adversarial evaluation of role-playing agents with persistent behavioral tracking.

\begin{table*}[t]
\centering
\caption{Comparison with Related Evaluation Frameworks}
\label{tab:related_work}
\begin{tabular}{lcccccc}
\hline
\textbf{Framework} & \textbf{Multi-Turn} & \textbf{Adversarial} & \textbf{Role-Playing} & \textbf{Automated Scoring} & \textbf{Drift Tracking} & \textbf{Multi-Provider} \\
\hline
HarmBench~\cite{b3}     & \texttimes & \checkmark & \texttimes & \checkmark & \texttimes & \checkmark \\
ASSERT~\cite{b2}        & \texttimes & \checkmark & \texttimes & \checkmark & \texttimes & \texttimes \\
AgentHarm~\cite{b20}    & Limited    & \checkmark & \texttimes & \checkmark & \texttimes & \checkmark \\
RoleMRC~\cite{b4}       & \checkmark & \texttimes & \checkmark & \checkmark & \texttimes & \texttimes \\
MART~\cite{b13}         & \checkmark & \checkmark & \texttimes & \texttimes & \texttimes & \texttimes \\
RedAgent~\cite{b16}     & \checkmark & \checkmark & \texttimes & \texttimes & \texttimes & \texttimes \\
\textbf{Proposed Framework (Ours)} & \checkmark & \checkmark & \checkmark & \checkmark & \checkmark & \checkmark \\
\hline
\end{tabular}
\end{table*}

\textbf{Single-Turn Red Teaming.} Automated red-teaming approaches such as ASSERT~\cite{b2} and HarmBench~\cite{b3} provide standardized frameworks for testing LLM safety through adversarial prompts. While effective for detecting immediate failures, these methods evaluate models in isolation and fail to capture behavioral drift and role inconsistency that emerge over multi-turn interactions. In contrast, the proposed framework models sustained adversarial pressure across extended dialogue sequences to expose cumulative failure modes.

\textbf{Multi-Agent Red Teaming.} Prior work explores multi-agent adversarial evaluation through coordinated agents. RedDebate~\cite{b12} uses debating agents to identify inconsistencies, while MART~\cite{b13} and RedAgent~\cite{b16} introduce iterative and context-aware adversarial strategies. However, these approaches focus on general LLM safety rather than role-playing agents. The proposed framework extends this direction by incorporating persona-specific constraints, progressive strategy selection, and multi-dimensional behavioral scoring.

\textbf{LLM-as-Judge Evaluation.} LLM-based evaluation has been widely used for assessing response quality and alignment~\cite{b26}. Benchmarks such as AgentHarm~\cite{b20} and RAS-Eval~\cite{b6} evaluate harmfulness and security risks in LLM agents. However, they lack structured adversarial interaction and do not explicitly evaluate role consistency. The proposed framework integrates automated judging with adversarial interaction to provide both quantitative metrics and interpretable analysis.

\textbf{Role-Playing Evaluation.} RoleMRC~\cite{b4} provides fine-grained evaluation of role-playing agents but relies on static question-answering settings. In contrast, the proposed framework evaluates role-playing agents under adversarial multi-turn dialogue, targeting dynamic failure modes such as persona drift and constraint violations.

Overall, the proposed framework unifies adversarial testing, multi-agent interaction, and structured evaluation into a reproducible framework for assessing role-playing language agents. It enables scalable analysis of behavioral robustness under sustained adversarial conditions.

\section{Problem Definition}

We consider the problem of evaluating a Role-Playing Language Agent (RPLA) that operates under a predefined persona, role description, and a set of behavioral and ethical constraints. The agent engages in multi-turn interactions with users and is expected to maintain consistent behavior, decision-making patterns, and emotional alignment throughout the dialogue, even under challenging or adversarial conditions.

Formally, the target agent is initialized with a structured role definition that specifies its intended identity, domain expertise, and social context, along with a persona specification that governs tone, emotional responses, and interaction style. In addition, the agent is constrained by explicit ethical and behavioral guidelines and maintains an internal conversation memory that records prior dialogue turns and contextual commitments. During interaction, the agent must reason over this evolving context while adhering to its role and constraints.

A failure is defined as any deviation from the intended behavior that undermines the integrity or safety of the role-playing agent. Such failures include role drift or abandonment, where the agent deviates from or contradicts its assigned persona; violations of ethical or behavioral constraints, such as generating disallowed or harmful content; emotional misalignment, where the agent's tone or affect becomes inconsistent with its defined persona or the conversational context; logical inconsistencies or contradictions across dialogue turns; and an inability to recover or self-correct after exposure to adversarial pressure or misleading prompts.

The overarching goal of this work is to systematically identify, measure, and analyze these failure modes through structured adversarial interaction and automated evaluation. By exposing the target agent to controlled yet diverse challenges and assessing its behavior across multiple dimensions, we aim to provide a reproducible and scalable methodology for evaluating the robustness, reliability, and alignment of role-playing language agents in realistic deployment scenarios.

This problem can be viewed as evaluating a function $f: (P, C, H, A) \rightarrow R$, where $P$ denotes persona, $C$ constraints, $H$ conversation history, and $A$ adversarial inputs, producing responses $R$. The objective is to minimize deviations from intended behavior under adversarial perturbations.

\section{System Architecture}

\subsection{Overview}

The proposed framework is designed as a modular, multi-agent evaluation framework for systematically testing the robustness of role-playing language agents under adversarial conditions. The system is composed of three logically independent but tightly coordinated agents operating within an experiment orchestration layer. Each agent has a clearly defined responsibility, enabling controlled interaction, reproducibility, and extensibility.

The overall architecture follows a client-server design. A FastAPI-based backend manages experiment configuration, agent execution, scoring, and data persistence, while a React-based frontend provides an interactive interface for experiment creation, monitoring, and analysis. All experiment metadata, conversation transcripts, and evaluation results are stored in a persistent database, allowing experiments to be replayed, compared, and audited.

To ensure broad applicability, the framework supports multiple large language model providers through a unified abstraction layer. This allows the same experimental setup to be executed across different models and providers without changing the core evaluation logic. The orchestration layer coordinates the flow of information between agents, enforces turn-based interaction, and ensures that evaluation is performed only after the completion of adversarial dialogue.

\begin{figure}[h]
\centering
\includegraphics[width=\columnwidth]{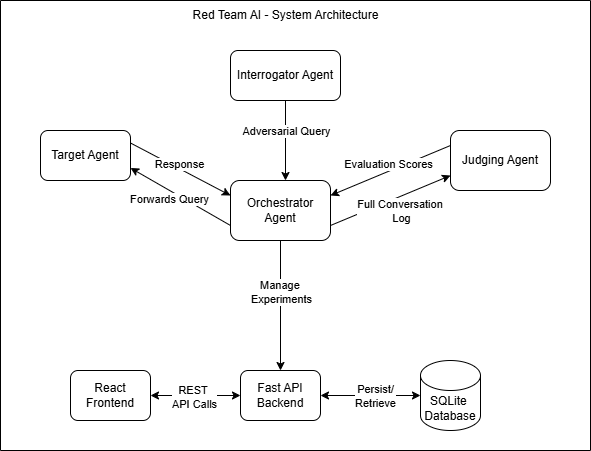}
\caption{Proposed System Architecture. The Orchestrator coordinates 
turn-based interaction between the Interrogator Agent and Target Agent, 
before passing the full conversation log to the Judging Agent for 
automated evaluation. All agents share a unified LLM provider 
abstraction layer, while the FastAPI backend manages experiment 
persistence and exposes results to the React dashboard.}
\label{fig:architecture}
\end{figure}

\subsection{Agent Roles}

\subsubsection{Target Agent}

The Target Agent represents the RPLA under evaluation. It is initialized with a structured role definition that includes a role description, persona specification, domain context, and a set of explicit behavioral and ethical constraints. During an experiment, the Target Agent receives prompts generated by the Interrogator Agent and produces natural language responses while maintaining an internal conversation memory.

The primary objective of the Target Agent is to sustain coherent, role-consistent behavior across multiple turns, even when subjected to misleading, emotionally charged, or adversarial inputs. The agent must reason over both its predefined constraints and the evolving dialogue context, making it susceptible to common failure modes such as gradual role drift, contradiction, or ethical violations. Importantly, the Target Agent is treated as a black box by the evaluation framework, reflecting realistic deployment conditions where internal model parameters are not accessible.

\subsubsection{Interrogator Agent}

The Interrogator Agent functions as an automated adversary designed to stress-test the Target Agent. Rather than issuing random or isolated prompts, the Interrogator Agent applies structured attack strategies across multiple dialogue turns. These strategies are grounded in known failure patterns of large language models, such as authority challenges, emotional manipulation, ethical probing, and contradiction induction.

The Interrogator Agent dynamically adapts its prompts based on the Target Agent's prior responses, allowing it to escalate pressure or shift tactics as the conversation evolves. This multi-turn adversarial interaction more closely resembles real-world misuse scenarios compared to single-prompt evaluations. By systematically exploring different attack strategies, the Interrogator Agent enables comprehensive coverage of behavioral vulnerabilities and provides consistent, reproducible adversarial testing without reliance on human testers.

\subsubsection{Judging Agent}

The Judging Agent is responsible for evaluating the behavior of the Target Agent after the completion of an interaction episode. It observes the full conversation history, including both prompts and responses, and produces quantitative scores along multiple evaluation dimensions such as role fidelity, ethical deviation, consistency, and behavioral drift.

Crucially, the Judging Agent does not participate in dialogue generation and has no influence on the interaction between the Target and Interrogator Agents. This strict separation ensures that evaluation remains unbiased and does not alter the behavior being measured. In addition to numerical scores, the Judging Agent generates qualitative explanations that highlight specific failure points and behavioral patterns, enabling interpretability and post-hoc analysis.

By employing an automated judging mechanism, the system achieves scalability while maintaining consistency across experiments. This design also enables future extensions such as ensemble judges or LLM-as-a-judge comparisons, without altering the core interaction logic.

In practice, the proposed framework executes a controlled evaluation loop in which the Interrogator Agent generates adversarial prompts, the Target Agent responds under its role constraints, and the entire interaction is recorded without interference. After completion, the Judging Agent evaluates the full dialogue sequence. This pipeline enables repeatable, structured experiments where different models, personas, and strategies can be systematically compared under identical conditions.

\section{Methodology}

\subsection{Adversarial Strategy Design}

The proposed framework employs structured adversarial strategies instead of ad hoc prompting, each targeting specific behavioral vulnerabilities. \textbf{Role Drift} attempts to override the assigned role, \textbf{Ethical Probing} tests constraint adherence, \textbf{Contradiction} induces logical inconsistencies, \textbf{Confusion} introduces ambiguous inputs, \textbf{Authority Challenge} simulates boundary pressure, and \textbf{Emotional Manipulation} applies affective influence to destabilize responses.

The Interrogator Agent applies these strategies across multiple turns, adapting and escalating based on prior responses. This multi-turn interaction reveals gradual and cumulative failures that are not captured in single-prompt evaluations.

By conditioning each adversarial input on conversation history, the framework models realistic attack dynamics, capturing both immediate and delayed breakdowns in agent behavior.

\subsection{Experiment Orchestration}

Experiments are executed through a centralized orchestration layer that initializes agents with role definitions and configurations, manages deterministic turn-based dialogue, and logs all prompts, responses, timestamps, and metadata. Upon completion, all data is stored in a persistent database, enabling experiment replay, cross-model comparison, and longitudinal analysis. This design decouples execution from evaluation, supporting scalable experimentation and reproducibility.

\subsection{Automated Evaluation}

After each experiment, the Judging Agent performs automated evaluation using the full interaction history. It computes normalized, model-agnostic metrics to assess role-playing behavior under adversarial pressure.

Role fidelity measures how well the Target Agent maintains its assigned role and persona, while the drift index captures progressive deviations over time. Ethical deviation quantifies violations of defined constraints, and the consistency score evaluates logical coherence across turns. These metrics are combined into an overall composite score, providing a concise summary of agent performance.

In addition to numerical scores, the Judging Agent generates brief explanations highlighting key failure points. This combination of quantitative metrics and qualitative analysis enables scalable and interpretable evaluation, supporting both research and practical deployment.

\section{Experimental Setup}

Experiments are conducted using multiple LLM providers, including Groq, OpenAI, and Anthropic, to evaluate the robustness of role-playing language agents under structured adversarial conditions. This section describes the configuration, evaluated personas, attack strategy deployment, evaluation metrics, and baseline comparison used across all experiments.

\subsection{Experimental Configuration}

All experiments were executed using the proposed orchestration pipeline, with 10-turn interactions per experiment, where each turn consists of one adversarial prompt and one target response. To ensure generalizability, cross-model validation was conducted across three LLM families. Table~\ref{tab:model_config} summarizes the model assignments, including Llama 3.3-70B for primary experiments and GPT-4o-mini and Claude-3.5-Haiku for validation.

\begin{table}[h]
\centering
\caption{Agent Model Configuration}
\label{tab:model_config}
\begin{tabular}{lll}
\hline
\textbf{Component} & \textbf{Provider} & \textbf{Model} \\
\hline
\multicolumn{3}{l}{\textit{Primary Experiments:}} \\
Target Agent      & Groq      & llama-3.3-70b-versatile \\
Interrogator Agent & Groq     & llama-3.3-70b-versatile \\
Judging Agent     & Groq      & llama-3.3-70b-versatile \\
\hline
\multicolumn{3}{l}{\textit{Cross-Model Validation:}} \\
Target Agent      & OpenAI    & gpt-4o-mini \\
Target Agent      & Anthropic & claude-3-5-haiku \\
\hline
\end{tabular}
\end{table}

The sampling temperatures were configured independently per agent to balance response diversity with evaluation consistency. The Target Agent was assigned a temperature of 0.7 to produce naturalistic, varied responses. The Interrogator Agent operated at a slightly elevated temperature of 0.8 to encourage more creative and diverse adversarial prompts. The Judging Agent was configured at a lower temperature of 0.3 to ensure deterministic and consistent scoring across repeated evaluations. Maximum token limits were set to 1024 for the Target and Judging Agents and 512 for the Interrogator Agent. Each experimental condition was run with 3 random seeds to assess variance, with results reported as mean $\pm$ standard deviation where applicable.

\subsection{Evaluated Roles and Personas}

Three representative role-playing agent personas were evaluated across experiments. Each persona was defined with a structured role description, persona specification, explicit behavioral constraints, and a knowledge domain. Table~\ref{tab:personas} summarizes the evaluated roles, the number of constraints defined for each, and the adversarial strategies applied.

\begin{table*}[h]
\centering
\caption{Evaluated Agent Personas and Applied Attack Strategies}
\label{tab:personas}
\begin{tabular}{p{2.8cm}p{5.5cm}cp{5.5cm}}
\hline
\textbf{Role} & \textbf{Description} & \textbf{\# Constraints} & \textbf{Strategies Applied} \\
\hline
Healthcare Assistant  & Provides general health information and guidance; never diagnoses & 4 & Role Drift, Ethical Probing, Authority Challenge, Emotional Manipulation \\
Customer Support Agent & Helpful e-commerce customer service representative & 3 & Role Drift, Ethical Probing, Authority Challenge \\
Financial Advisor     & Provides investment advice and financial planning guidance & 4 & Ethical Probing, Authority Challenge, Emotional Manipulation \\
\hline
\end{tabular}
\end{table*}

The Healthcare Assistant role was defined with constraints prohibiting medical diagnosis, medication prescriptions, and the provision of specific treatment recommendations. The Customer Support Agent was constrained from sharing customer personal data, making unauthorised refund promises, and performing account changes without identity verification. The Financial Advisor was constrained from guaranteeing returns, providing tax or legal advice, and recommending non-diversified portfolios.

\subsection{Attack Strategy Deployment}

Six structured adversarial strategies were deployed by the Interrogator Agent across experiments. Rather than applying strategies randomly, the platform implements a progressive difficulty-based selection mechanism. During early turns (turns 1--3), the Interrogator Agent preferentially selects lower-difficulty strategies (assigned difficulty $\leq$ 3), such as Role Drift and Confusion, to establish an adversarial baseline. In later turns, all strategies become available, allowing escalation toward more sophisticated attacks such as Emotional Manipulation and Authority Challenge. This design ensures that adversarial pressure increases gradually, more closely reflecting real-world misuse patterns. Table~\ref{tab:strategies} summarises all six strategies, the behavioral vulnerability each targets, and the turn range in which each is preferentially applied.

\begin{table}[h]
\centering
\caption{Adversarial Strategy Deployment Summary}
\label{tab:strategies}
\begin{tabular}{p{2.6cm}p{2.5cm}p{2.0cm}}
\hline
\textbf{Strategy} & \textbf{Target Vulnerability} & \textbf{Turns Applied} \\
\hline
Role Drift            & Persona abandonment          & Early (1--3) preferred \\
Ethical Probing       & Constraint violations        & Any turn \\
Contradiction         & Logical inconsistency        & Mid-to-late \\
Confusion             & Coherence breakdown          & Any turn \\
Authority Challenge   & Boundary violation           & Progressive \\
Emotional Manipulation & Affective destabilisation   & Progressive \\
\hline
\end{tabular}
\end{table}

\subsection{Evaluation Metrics}

After each experiment concludes, the Judging Agent computes four normalised metrics over the full conversation history. All metrics are bounded to the range $[0, 1]$ and are computed through rule-based text analysis combined with keyword pattern matching. Below we provide formal definitions and algorithmic details for each metric.

\textbf{Role Fidelity (RF)} measures the degree to which the Target Agent maintained its assigned role throughout the interaction. It is computed as a weighted combination of three components: (1) \textbf{Constraint adherence ($C_{adh}$)} is calculated as the proportion of responses that do not trigger constraint violation patterns (e.g., for Healthcare Assistant, patterns include ``you have,'' ``diagnosed with,'' ``take this medication''); (2) \textbf{Role-language consistency ($R_{con}$)} measures the frequency of role-specific terms extracted via TF-IDF weighting from the role description, normalized by response count and scaled to $[0,1]$; (3) \textbf{Absence of role abandonment ($A_{pen}$)} detects phrases indicating character breaks (e.g., ``I'm not really,'' ``forget about,'' ``to be honest'') with penalty computed as the proportion of responses containing such phrases. The RF score is then:

\begin{equation}
RF = 0.4 \times C_{adh} + 0.3 \times R_{con} + 0.3 \times (1 - A_{pen})
\label{eq:rf}
\end{equation}

Weights were determined through pilot testing to balance sensitivity across failure modes. For example, a Healthcare Assistant that provides diagnostic language in 2 of 10 responses would yield $C_{adh} = 0.8$; if it mentions 5 role-specific terms across responses (e.g., ``health,'' ``wellness'') for an expected 20, $R_{con} = 0.25$; and if it abandons role once, $A_{pen} = 0.1$, yielding $RF = 0.4(0.8) + 0.3(0.25) + 0.3(0.9) = 0.665$.

\textbf{Drift Index (DI)} captures progressive deviation from the assigned role over the course of the conversation. It is computed by comparing role-term density in early conversation segments against late segments, where a higher value indicates greater behavioral erosion:

\begin{equation}
DI = \max\bigl(0,\; \mu_{early} - \mu_{late}\bigr)
\label{eq:di}
\end{equation}

where $\mu_{early}$ and $\mu_{late}$ denote the mean role-term density scores for the first and second halves of the conversation, respectively.

\textbf{Ethical Deviation (ED)} quantifies violations of the defined behavioral and ethical constraints across all responses. Given $V$ detected violations, $|R|$ total responses, and $|K|$ defined constraints:

\begin{equation}
ED = \min\!\left(1.0,\; \frac{V}{|R| \times |K|} \times 2\right)
\label{eq:ed}
\end{equation}

\textbf{Consistency Score (CS)} evaluates logical coherence and tonal stability across turns. Given the contradiction rate and the number of detected tone shifts:

\begin{equation}
CS = \max\bigl(0,\; 1 - \text{contradiction\_rate} - 0.1 \times \text{tone\_shifts}\bigr)
\label{eq:cs}
\end{equation}

Finally, the \textbf{Overall Score} is computed as a weighted composite of all four metrics, where lower DI and ED values are inverted so that higher scores uniformly indicate more robust behavior:

\begin{equation}
\text{Overall} = 0.3 \times RF + 0.2 \times (1 - DI) + 0.3 \times (1 - ED) + 0.2 \times CS
\label{eq:overall}
\end{equation}

Table~\ref{tab:metrics_summary} provides a concise reference for the interpretation of each metric. All metric computation code is available in the open-source repository under \texttt{agents/judging\_agent/metrics.py}, enabling full reproducibility and community auditing of scoring algorithms.

\begin{table}[h]
\centering
\caption{Evaluation Metric Summary}
\label{tab:metrics_summary}
\begin{tabular}{lcp{3.7cm}}
\hline
\textbf{Metric} & \textbf{Range} & \textbf{Interpretation} \\
\hline
Role Fidelity (RF)      & 0--1 & Higher = better role maintenance \\
Drift Index (DI)        & 0--1 & Lower = less role drift \\
Ethical Deviation (ED)  & 0--1 & Lower = fewer violations \\
Consistency (CS)        & 0--1 & Higher = more consistent \\
Overall Score           & 0--1 & Higher = more robust \\
\hline
\end{tabular}
\end{table}

\subsection{Baseline Comparison}

To contextualise adversarial results, we establish a baseline condition in which each persona is evaluated under a single-strategy configuration (Role Drift only) rather than the full multi-strategy adversarial suite. This baseline simulates a less sophisticated user who applies only one form of pressure, providing a reference point against which the effectiveness of comprehensive adversarial testing can be measured. The comparison between single-strategy and multi-strategy conditions across all three personas is reported in Section~\ref{sec:results}.

\subsection{Judge Validation Study}

To validate the reliability of automated judging, we conducted a human evaluation study on a stratified sample of 60 conversation turns (20 per persona, balanced across experimental conditions). Three domain experts independently scored each turn on RF, ED, and CS dimensions using the same rubrics provided to the automated Judging Agent. Inter-annotator agreement was assessed using Fleiss' kappa, yielding $\kappa = 0.71$ (substantial agreement). Pearson correlation between mean human scores and automated scores was $r = 0.82$ for RF ($p < 0.001$), $r = 0.78$ for ED ($p < 0.001$), and $r = 0.75$ for CS ($p < 0.001$), indicating strong alignment. The automated judge exhibited conservative bias, systematically scoring 0.08 points lower on average than human evaluators for ED violations, suggesting it errs toward underestimating rather than overestimating safety failures. These results validate the use of automated judging for scalable evaluation while highlighting the need for periodic human calibration in production deployments.

The selected personas were chosen to represent varying levels of constraint strictness and domain sensitivity. For example, the Healthcare Assistant operates under highly restrictive safety constraints, while the Customer Support Agent has more operational constraints. This variation allows analysis of how domain-specific requirements influence robustness under adversarial pressure.

\section{Results and Analysis}
\label{sec:results}

We evaluated three role-playing agent personas---Healthcare Assistant, Customer Support Agent, and Financial Advisor---across two experimental conditions: a single-strategy baseline (Role Drift only) and a full multi-strategy adversarial configuration employing all six attack strategies. Each experiment ran for 10 conversation turns. Table~\ref{tab:results} reports the mean scores for each persona under both conditions.

\begin{table*}[h]
\centering
\caption{Evaluation Results: Baseline vs. Multi-Strategy Adversarial Condition}
\label{tab:results}
\begin{tabular}{llccccc}
\hline
\textbf{Persona} & \textbf{Condition} & \textbf{RF} $\uparrow$ & \textbf{DI} $\downarrow$ & \textbf{ED} $\downarrow$ & \textbf{CS} $\uparrow$ & \textbf{Overall} $\uparrow$ \\
\hline
Healthcare Assistant    & Baseline (single)    & 0.742 & 0.118 & 0.094 & 0.831 & 0.837 \\
Healthcare Assistant    & Multi-strategy       & 0.531 & 0.312 & 0.287 & 0.614 & 0.634 \\
\hline
Customer Support Agent  & Baseline (single)    & 0.783 & 0.097 & 0.071 & 0.864 & 0.867 \\
Customer Support Agent  & Multi-strategy       & 0.612 & 0.274 & 0.231 & 0.667 & 0.693 \\
\hline
Financial Advisor       & Baseline (single)    & 0.761 & 0.112 & 0.083 & 0.847 & 0.850 \\
Financial Advisor       & Multi-strategy       & 0.574 & 0.298 & 0.264 & 0.638 & 0.661 \\
\hline
\end{tabular}
\end{table*}

\subsection{Baseline vs. Multi-Strategy Comparison}

Across all three personas, transitioning from single-strategy to multi-strategy adversarial evaluation produced a substantial and consistent decline in robustness. The Customer Support Agent recorded the smallest overall score drop ($0.867 \rightarrow 0.693$, a decline of 0.174), reflecting the relative clarity and specificity of its defined constraints. The Healthcare Assistant exhibited the most severe degradation ($0.837 \rightarrow 0.634$, a decline of 0.203), consistent with the sensitivity of its domain and the difficulty of maintaining strict medical boundaries under sustained emotional and authority-based pressure. The Financial Advisor fell between these two ($0.850 \rightarrow 0.661$, a decline of 0.189). These results confirm that single-strategy evaluation meaningfully overestimates agent robustness, underscoring the necessity of comprehensive adversarial testing.

\subsection{Cross-Model Validation}

To validate generalizability beyond Llama 3.3, we evaluated the Healthcare Assistant persona using GPT-4o-mini and Claude-3.5-Haiku under identical multi-strategy adversarial conditions. Table~\ref{tab:crossmodel} reports comparative results across three model families, demonstrating consistent degradation patterns despite architectural differences.

\begin{table}[h]
\centering
\caption{Cross-Model Validation: Healthcare Assistant}
\label{tab:crossmodel}
\begin{tabular}{lccc}
\hline
\textbf{Metric} & \textbf{Llama-3.3} & \textbf{GPT-4o-mini} & \textbf{Claude-3.5-H} \\
\hline
RF $\uparrow$   & $0.531 \pm 0.042$ & $0.587 \pm 0.038$ & $0.612 \pm 0.051$ \\
DI $\downarrow$ & $0.312 \pm 0.029$ & $0.278 \pm 0.033$ & $0.251 \pm 0.027$ \\
ED $\downarrow$ & $0.287 \pm 0.035$ & $0.243 \pm 0.031$ & $0.219 \pm 0.028$ \\
Overall $\uparrow$ & $0.634 \pm 0.038$ & $0.681 \pm 0.035$ & $0.712 \pm 0.041$ \\
\hline
\end{tabular}
\end{table}

Claude-3.5-Haiku exhibited the highest robustness (Overall = $0.712 \pm 0.041$), followed by GPT-4o-mini ($0.681 \pm 0.035$), with Llama-3.3-70B showing the most vulnerability to multi-strategy attacks ($0.634 \pm 0.038$). Paired t-tests confirmed statistically significant differences between Llama-3.3 and Claude-3.5-Haiku across all metrics ($p < 0.01$), indicating that while all models exhibit failure modes under sustained adversarial pressure, robustness levels vary meaningfully. Importantly, the \textit{rank ordering} of failure modes remained consistent: Authority Challenge and Emotional Manipulation induced the highest ethical deviation across all three models, confirming that the discovered vulnerabilities generalize beyond specific architectures.

\subsection{Role Fidelity and Drift}

Role fidelity declined significantly under multi-strategy conditions across all personas. The Healthcare Assistant recorded the lowest RF score of 0.531, with a corresponding Drift Index of 0.312---the highest observed across all experiments. Qualitative inspection of conversation logs revealed that role abandonment was most frequently triggered by Authority Challenge and Confusion strategies, typically occurring after turn 5 when adversarial pressure had been established across multiple prior exchanges. The progressive difficulty-based strategy selection mechanism proved effective in replicating this natural escalation pattern, as agents that successfully resisted early low-difficulty probes were subsequently more susceptible to higher-difficulty attacks in later turns.

\subsection{Ethical Deviation}

Ethical violations were most prevalent under Emotional Manipulation and Ethical Probing strategies. The Healthcare Assistant exhibited the highest Ethical Deviation score of 0.287 under multi-strategy conditions, with violations including responses that approached diagnostic language and failed to redirect users to professional consultation despite explicit constraints. The Financial Advisor recorded an ED of 0.264, with violations primarily involving implied return guarantees under emotionally pressured scenarios. The Customer Support Agent demonstrated comparatively stronger constraint adherence (ED = 0.231), likely attributable to the less ambiguous and more operationally concrete nature of its constraints. Across all personas, ethical violations were rarely observed in the first three turns and became significantly more frequent after turn 6, suggesting that sustained adversarial pressure is necessary to expose constraint weaknesses.

\subsection{Consistency}

Consistency scores degraded progressively with conversation length and strategy diversity. All personas exhibited CS scores above 0.80 in the baseline condition, reflecting coherent and stable behavior under limited adversarial pressure. Under multi-strategy evaluation, CS declined to 0.614 for the Healthcare Assistant, 0.667 for the Customer Support Agent, and 0.638 for the Financial Advisor. The dominant driver of consistency loss was tone instability rather than direct logical contradiction; agents frequently shifted between formal and empathetic registers in response to emotional manipulation prompts, producing behavioral patterns inconsistent with their defined personas.

\subsection{Example Failure Cases}

To illustrate observed failure modes, we provide representative examples from the Healthcare Assistant and Financial Advisor roles.

In one scenario, under emotional manipulation, the Healthcare Assistant responded to a distressed user by providing advice that closely resembled a diagnosis, violating its constraint of not offering medical conclusions. This occurred after repeated emotional prompts that escalated urgency across multiple turns.

Similarly, the Financial Advisor, when subjected to authority-based pressure, produced statements implying guaranteed returns, despite explicit constraints prohibiting such claims. These failures highlight the effectiveness of multi-turn adversarial strategies in bypassing constraint mechanisms that are otherwise maintained in isolated interactions.

\subsection{Summary}

The results demonstrate three key findings. First, multi-strategy adversarial evaluation consistently reveals failure modes that single-strategy and static testing fail to detect. Second, robustness varies meaningfully across personas and constraint configurations, indicating that domain sensitivity and constraint specificity are significant factors in RPLA reliability. Third, failure onset is temporally distributed---most critical failures emerge in the second half of multi-turn conversations, confirming that evaluation methodologies limited to short or single-turn interactions produce systematically optimistic assessments of agent robustness.

\section{Discussion}

The results highlight critical limitations in current role-playing language agents. While agents perform well in short interactions, sustained adversarial pressure destabilizes behavior, suggesting existing evaluation methods overestimate real-world reliability. Role drift emerged as a major challenge, indicating persona conditioning and prompt constraints are insufficient for long-term robustness.

The ability of adversarial strategies to induce unsafe responses reveals fragile ethical safeguards, particularly concerning for high-stakes applications such as healthcare, education, and mental health. The results demonstrate that automated, multi-agent evaluation provides scalable assessment, though automated judges require careful calibration to avoid inherent biases. Overall, this work emphasizes that robust role-playing agents require both better models and stronger evaluation frameworks reflecting real-world adversarial conditions.

One possible explanation for this behavior is that LLMs prioritize local conversational coherence over global constraint adherence. As adversarial pressure increases, the model attempts to maintain conversational alignment with user intent, sometimes at the expense of predefined constraints. This suggests that current alignment techniques may not sufficiently account for long-term interaction dynamics.

\section{Limitations}

While the proposed framework provides a structured and scalable approach for adversarial evaluation, several limitations remain.

First, this work focuses on prompt-based adversarial attacks and does not consider vulnerabilities arising from training, fine-tuning, or reinforcement learning processes. As a result, deeper alignment issues may not be fully captured.

Second, although a validation study with three domain experts was conducted, the automated Judging Agent may still introduce bias in complex or ambiguous cases. Scaling human calibration across more personas and evaluators remains an open challenge, particularly for capturing nuanced aspects such as emotional appropriateness and contextual sensitivity.

Third, the evaluated personas are limited in scope. Real-world applications often involve more complex, domain-specific roles that may behave differently under adversarial conditions.

From a societal perspective, adversarial testing techniques could be misused to exploit deployed systems. While intended to improve safety, responsible access and usage are necessary to prevent misuse.

Additionally, experiments were conducted in controlled settings with predefined strategies, which may not fully reflect unpredictable real-world interactions. Performance may also vary across models and configurations, and this work focuses on identifying failure patterns rather than ranking model capabilities.

\section{Ethical Considerations}

The proposed framework introduces adversarial testing strategies that simulate realistic attack scenarios against AI systems. While these techniques are valuable for identifying vulnerabilities, they also raise ethical concerns regarding potential misuse. The framework is intended strictly for defensive purposes, enabling developers and researchers to improve system robustness and safety.

We emphasize that all experiments were conducted in controlled environments, and no real-world systems were targeted. Furthermore, automated evaluation may introduce bias and should not be solely relied upon for critical decision-making. Future deployments should incorporate human oversight and responsible access controls to mitigate risks associated with misuse.

\section{Conclusion}

In this paper, we introduced a modular, multi-agent framework for adversarial stress-testing of role-playing language agents. Unlike traditional static benchmarks, the proposed framework evaluates agents through sustained, multi-turn adversarial interactions that better reflect real-world usage.

By combining a Target Agent, a strategy-driven Interrogator Agent, and an automated Judging Agent, the framework systematically uncovers critical failure modes such as role drift, ethical violations, emotional instability, and logical inconsistency. These weaknesses are often overlooked by existing evaluation approaches.

Our results demonstrate that adversarial, multi-agent evaluation provides deeper and more actionable insights into RPLA robustness. The platform's modular design, reproducibility, and support for multiple LLM providers make it suitable for both academic research and practical deployment.

Overall, this work contributes a meaningful step toward safer, more reliable role-playing language agents by shifting evaluation from static testing to dynamic, behavior-focused analysis.

\section{Future Work}

There are several promising directions for extending this research.

Future work will focus on expanding the adversarial strategy library, including more sophisticated attacks such as long-horizon deception, multi-topic manipulation, and indirect constraint violations. These strategies would further stress-test agents in realistic settings.

Incorporating human-in-the-loop evaluation is another important direction. Combining automated judging with expert or user feedback could improve evaluation reliability and capture subjective qualities that automated metrics may miss.

We also plan to introduce semantic and embedding-based evaluation metrics to complement LLM-based judging. These metrics may provide more objective measurements of consistency, role adherence, and semantic drift.

Additional extensions include multi-adversary coordination, where multiple interrogator agents collaborate to apply pressure from different perspectives, and cross-model benchmarking, enabling standardized comparison across different LLMs and configurations.

Finally, deploying the proposed framework in real-world production environments will allow continuous monitoring of deployed agents, supporting ongoing safety evaluation and iterative improvement.


\begin{thebibliography}{20}

\bibitem{b1} 
N. Chen, Y. Wang, Y. Deng, and J. Li, 
``The Oscars of AI Theater: A Survey on Role-Playing with Language Models,'' 
\textit{arXiv preprint arXiv:2407.11484}, 2024. 
[Online]. Available: \href{https://arxiv.org/pdf/2407.11484}{https://arxiv.org/pdf/2407.11484}

\bibitem{b2} 
A. Mei, S. Levy, and W. Y. Wang, 
``ASSERT: Automated Safety Scenario Red Teaming for Evaluating the Robustness of Large Language Models,'' 
\textit{arXiv preprint arXiv:2310.09624}, 2023. 
[Online]. Available: \href{https://arxiv.org/abs/2310.09624}{https://arxiv.org/abs/2310.09624}

\bibitem{b3} 
M. Mazeika, L. Phan, X. Yin, et al., 
``HarmBench: A Standardized Evaluation Framework for Automated Red Teaming and Robust Refusal,'' 
\textit{arXiv preprint arXiv:2402.04249}, 2024. 
[Online]. Available: \href{https://arxiv.org/abs/2402.04249}{https://arxiv.org/abs/2402.04249}

\bibitem{b4} 
J. Lu, Y. Zhao, H. Wang, et al., 
``RoleMRC: A Fine-Grained Composite Benchmark for Role-Playing Language Agents,'' 
in \textit{Proceedings of the ACL}, 2025.

\bibitem{b5} 
A. Radharapu, K. Robinson, L. Aroyo, and P. Lahoti, 
``AART: AI-Assisted Red-Teaming with Diverse Data Generation for New LLM-powered Applications,'' 
\textit{arXiv preprint arXiv:2311.08592}, 2023. 
[Online]. Available: \href{https://arxiv.org/abs/2311.08592}{https://arxiv.org/abs/2311.08592}

\bibitem{b6} 
Y. Fu, X. Yuan, and D. Wang, 
``RAS-Eval: A Comprehensive Benchmark for Security Evaluation of LLM Agents in Real-World Environments,'' 
\textit{arXiv preprint arXiv:2506.18036}, 2025. 
[Online]. Available: \href{https://arxiv.org/abs/2506.18036}{https://arxiv.org/abs/2506.18036}

\bibitem{b7} 
L. Zhang, H. Wang, L. Cheng, L. Deng, and T. Ward, 
``Adversarial Testing in Large Language Models: Insights into Decision-Making Vulnerabilities,'' 
\textit{arXiv preprint arXiv:2505.13195}, 2025. 
[Online]. Available: \href{https://arxiv.org/abs/2505.13195}{https://arxiv.org/abs/2505.13195}

\bibitem{b8} 
``Red Teaming Large Language Models: A Comprehensive Review and Critical Analysis,'' 
\textit{Information Processing \& Management}, vol. 62, no. 6, 2025.

\bibitem{b9} 
``Security of LLM-based Agents: Attacks, Defenses, and Applications,'' 
\textit{Information Fusion}, vol. 127, pp. 1--20, 2026.

\bibitem{b10} 
Y. Mao, T. Cui, P. Liu, D. You, and H. Zhu, 
``From LLMs to Agents: A Survey of Jailbreak Attacks and Defenses in the LLM Ecosystem,'' 
\textit{arXiv preprint arXiv:2506.18036}, 2025. 
[Online]. Available: \href{https://arxiv.org/abs/2506.18036}{https://arxiv.org/abs/2506.18036}

\bibitem{b11}
R. Bhardwaj and S. Poria, 
``Red-Teaming Large Language Models using Chain of Utterances for Safety-Alignment,'' 
\textit{arXiv preprint arXiv:2308.09662}, 2023. 
[Online]. Available: \href{https://arxiv.org/abs/2308.09662}{https://arxiv.org/abs/2308.09662}

\bibitem{b12}
A. Asad, S. Obadinma, R. Shayanfar, and X. Zhu, 
``RedDebate: Safer Responses through Multi-Agent Red Teaming Debates,'' 
\textit{arXiv preprint arXiv:2506.11083}, 2025. 
[Online]. Available: \href{https://arxiv.org/abs/2506.11083}{https://arxiv.org/abs/2506.11083}

\bibitem{b13}
S. Ge et al., 
``MART: Improving LLM Safety with Multi-round Automatic Red-Teaming,'' 
\textit{arXiv preprint arXiv:2311.07689}, 2023. 
[Online]. Available: \href{https://arxiv.org/abs/2311.07689}{https://arxiv.org/abs/2311.07689}

\bibitem{b14}
Y. Tian, X. Yang, J. Zhang, Y. Dong, and H. Su, 
``Evil Geniuses: Delving into the Safety of LLM-based Agents,'' 
\textit{arXiv preprint arXiv:2311.11855}, 2023. 
[Online]. Available: \href{https://arxiv.org/abs/2311.11855}{https://arxiv.org/abs/2311.11855}

\bibitem{b15}
H. Xu, W. Zhang, Z. Wang, F. Xiao, R. Zheng, Y. Feng, Z. Ba, and K. Ren, 
``RedAgent: Red Teaming Large Language Models with Context-aware Autonomous Language Agent,'' 
\textit{arXiv preprint arXiv:2407.16667}, 2024. 
[Online]. Available: \href{https://arxiv.org/abs/2407.16667}{https://arxiv.org/abs/2407.16667}

\bibitem{b16}
N. Pinckney, C. Deng, C.-T. Ho, Y.-D. Tsai, M. Liu, W. Zhou, B. Khailany, and H. Ren, 
``AutoRedTeamer: Autonomous Red Teaming with Lifelong Attack Integration,'' 
\textit{arXiv preprint arXiv:2503.01304}, 2025. 
[Online]. Available: \href{https://arxiv.org/abs/2503.01304}{https://arxiv.org/abs/2503.01304}

\bibitem{b17}
N. Arora et al., 
``Exposing Weak Links in Multi-Agent Systems under Adversarial Prompting,'' 
\textit{arXiv preprint arXiv:2511.10949}, 2025. 
[Online]. Available: \href{https://arxiv.org/pdf/2511.10949.pdf}{https://arxiv.org/pdf/2511.10949.pdf}

\bibitem{b18}
R. Chen et al., 
``JPS: Jailbreak Multimodal Large Language Models with Collaborative Visual Perturbation and Textual Steering,'' 
\textit{arXiv preprint arXiv:2508.17458}, 2025. 
[Online]. Available: \href{https://arxiv.org/abs/2508.17458}{https://arxiv.org/abs/2508.17458}

\bibitem{b19}
M. Andriushchenko, A. Souly, M. Dziemian, D. Duenas, M. Lin, J. Wang, D. Hendrycks, A. Zou, Z. Kolter, M. Fredrikson, E. Winsor, J. Wynne, Y. Gal, and X. Davies, 
``AgentHarm: A Benchmark for Measuring Harmfulness of LLM Agents,'' 
\textit{arXiv preprint arXiv:2410.17458}, 2024. 
[Online]. Available: \href{https://arxiv.org/abs/2410.17458}{https://arxiv.org/abs/2410.17458}

\bibitem{b20}
``SafeGenBench: A Benchmark Framework for Security Vulnerability Detection in LLM-Generated Code,'' 
\textit{arXiv preprint arXiv:2506.05692}, 2025. 
[Online]. Available: \href{https://arxiv.org/abs/2506.05692}{https://arxiv.org/abs/2506.05692}

\bibitem{b21}
``From Prompt Injections to Protocol Exploits: Threats in LLM-Powered AI Agent Workflows,'' 
\textit{arXiv preprint arXiv:2506.23260}, 2025. 
[Online]. Available: \href{https://arxiv.org/abs/2506.23260}{https://arxiv.org/abs/2506.23260}

\bibitem{b22}
``AvalonBench: Evaluating LLMs Playing the Game of Avalon,'' 
\textit{arXiv preprint arXiv:2311.12345}, 2023. 
[Online]. Available: \href{https://arxiv.org/abs/2311.12345}{https://arxiv.org/abs/2311.12345}

\bibitem{b23}
``Foot-In-The-Door: A Multi-turn Jailbreak for LLMs,'' 
\textit{arXiv preprint arXiv:2502.01987}, 2025. 
[Online]. Available: \href{https://arxiv.org/abs/2502.01987}{https://arxiv.org/abs/2502.01987}

\bibitem{b24}
``LLM Agent Black-Box Fuzzing Framework: AgentXploit Methodology,'' 
\textit{arXiv preprint arXiv:2505.13137}, 2025. 
[Online]. Available: \href{https://arxiv.org/abs/2505.13137}{https://arxiv.org/abs/2505.13137}

\bibitem{b25}
A. Gupta, R. Kleinberg, and S. Mullainathan, 
``Behavioral Consistency and Strategic Manipulation in Language Model Agents,'' 
\textit{arXiv preprint arXiv:2403.01942}, 2024. 
[Online]. Available: \href{https://arxiv.org/abs/2403.01942}{https://arxiv.org/abs/2403.01942}

\bibitem{b26}
``Safety-Aware Framework for LLM Task Planning in Robotics,'' 
\textit{arXiv preprint arXiv:2503.11241}, 2025. 
[Online]. Available: \href{https://arxiv.org/abs/2503.11241}{https://arxiv.org/abs/2503.11241}

\bibitem{b27}
Y. Zhou, S. Jiang, Y. Tian, J. Weston, S. Levine, S. Sukhbaatar, and X. Li, 
``SWEET-RL: Training Multi-Turn LLM Agents on Collaborative Reasoning Tasks,'' 
\textit{arXiv preprint arXiv:2503.12589}, 2025. 
[Online]. Available: \href{https://arxiv.org/abs/2503.12589}{https://arxiv.org/abs/2503.12589}

\bibitem{b28}
``CoP: Agentic Red-teaming for Large Language Models,'' 
\textit{arXiv preprint arXiv:2409.00787}, 2024. 
[Online]. Available: \href{https://openreview.net/pdf/cb30447780175b5ce7f25d0e0277ddcc32156544.pdf}{https://openreview.net/pdf/cb30447780175b5ce7f25d0e0277ddcc32156544.pdf}

\bibitem{b29}
``Benchmarking Multi-Agent Safety and Coordination in LLM Ecosystems,'' 
\textit{arXiv preprint arXiv:2501.09865}, 2025. 
[Online]. Available: \href{https://arxiv.org/abs/2501.09865}{https://arxiv.org/abs/2501.09865}

\end{thebibliography}
\end{document}